\documentclass{article}

\usepackage[preprint, nonatbib]{neurips_2023}

\usepackage[utf8]{inputenc} 
\usepackage[T1]{fontenc}    
\usepackage{hyperref}       
\usepackage{url}            
\usepackage{booktabs}       
\usepackage{amsfonts}       
\usepackage{nicefrac}       
\usepackage{microtype}      
\usepackage{graphicx}
\usepackage{amsmath}
\usepackage[dvipsnames]{xcolor}
\usepackage{babel,blindtext}
\usepackage{algpseudocode}
\usepackage{algorithm}
\title{Priority Sampling of Large Language Models for Compilers}

\author{%
  Dejan ~Grubisic \thanks{Corresponding author: dx4@rice.edu} \\
  Rice University, Meta AI \\
  \And
  Chris Cummins \\
  Meta AI \\
  \And
  Volker Seeker \\
  Meta AI \\
  \And
  Hugh Leather \\
  Meta AI \\
}
  
\begin{document}

\maketitle

\begin{abstract}
Large language models show great potential in generating and optimizing code. Widely used sampling methods such as Nucleus Sampling increase the diversity of generation but often produce repeated samples for low temperatures and incoherent samples for high temperatures. Furthermore, the temperature coefficient has to be tuned for each task, limiting its usability. 
We present \emph{Priority Sampling}, a simple and deterministic sampling technique that produces unique samples ordered by the model's confidence. Each new sample expands the unexpanded token with the highest probability in the augmented search tree. Additionally, Priority Sampling supports generation based on regular expression that provides a controllable and structured exploration process.
Priority Sampling outperforms Nucleus Sampling for any number of samples, boosting the performance of the original model from 2.87\% to 5\% improvement over -Oz. Moreover, it outperforms the autotuner used for the generation of labels for the training of the original model in just 30 samples.
\end{abstract}

\section{Introduction and Motivation}
Large language models (LLMs) have proven their ability in the software engineering domain to generate the code and documentation~\cite{li_starcoder_2023, allal2023santacoder}, translate code between programming languages~\cite{transcoder, armengol2021learning}, write unit-tests~\cite{titanfuzz, schäfer2023adaptive}, as well as detect and fix bugs~\cite{ahmad2023fixing, xia2023automated}. CodeLlama~\cite{llama-code}, ChatGPT~\cite{openai2023gpt4}, and Codex~\cite{chen_evaluating_2021_codex} excel in various code manipulation tasks significantly improving the coding experience. Some of the models such as AlphaCode~\cite{li_competition-level_2022_alphacode} are pretrained on competitive programming tasks which enables the model to optimize code on the source level for several languages.

All these models could boost LLM's performance by generating an ensemble of diverse solutions, from which we evaluate and choose the best. This is usually done by increasing the entropy of generation~\cite{ravfogel2023nucleus_topp, li2020topk}, or expanding the search tree~\cite{xie2023beam, kool2019stochastic, vilnis2023arithmetic, shi2020incremental}.
Sampling also enables us to better understand the capacity of the LLM on a given task and the range of possible solutions. This is particularly important in code generation where generating a variety of responses can be valuable in exploring different implementation ideas.

Current sampling approaches have few major problems. Temperature-based sampling~\cite{ravfogel2023nucleus_topp, li2020topk} requires a significant amount of computation to find the optimal temperature. The optimal temperature may also depend on the context, which requires additional evaluations. Once we set the temperature, sampling often produces a large number of duplicates and semantically meaningless answers, wasting available samples.

\begin{figure}[t]
  \centering
  \includegraphics[width=0.7\columnwidth]{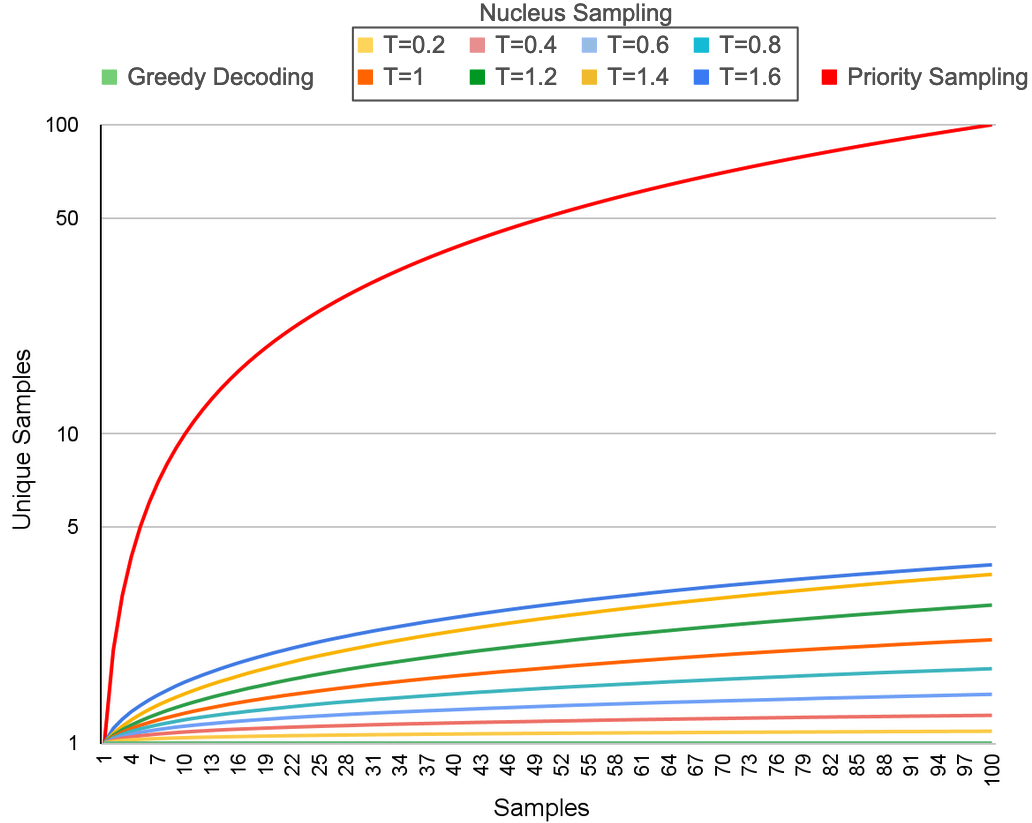}
  \caption{Average number of unique samples generated from 50k unseen test programs. Priority sampling produces a higher ratio of unique samples than nucleus sampling.}
  \label{fig:unique_samples}
\end{figure}

To motivate our approach, we show the average number of unique samples generated by Nucleus Sampling compared to Priority Sampling (Figure \ref{fig:unique_samples}). On 50k test examples, Nucleus Sampling generates less than 5 unique examples on average for 100 samples, while Priority Sampling generates 100. The reason for this is that Nucleus Sampling either chooses the high probability generation repeatedly or it outputs non-coherent output which we don't count as a meaningful unique sample.

In this work, we present Priority Sampling, a deterministic sampling technique that guarantees unique samples, for which the model has the highest confidence.  
Furthermore, we guarantee that produced samples will adhere to the regular expression (inspired by Willard~\cite{willard2023regex_sampling}), which is particularly important for code generation and code optimization.

\begin{figure}[h]
  \centering
  \includegraphics[width=0.77\columnwidth]{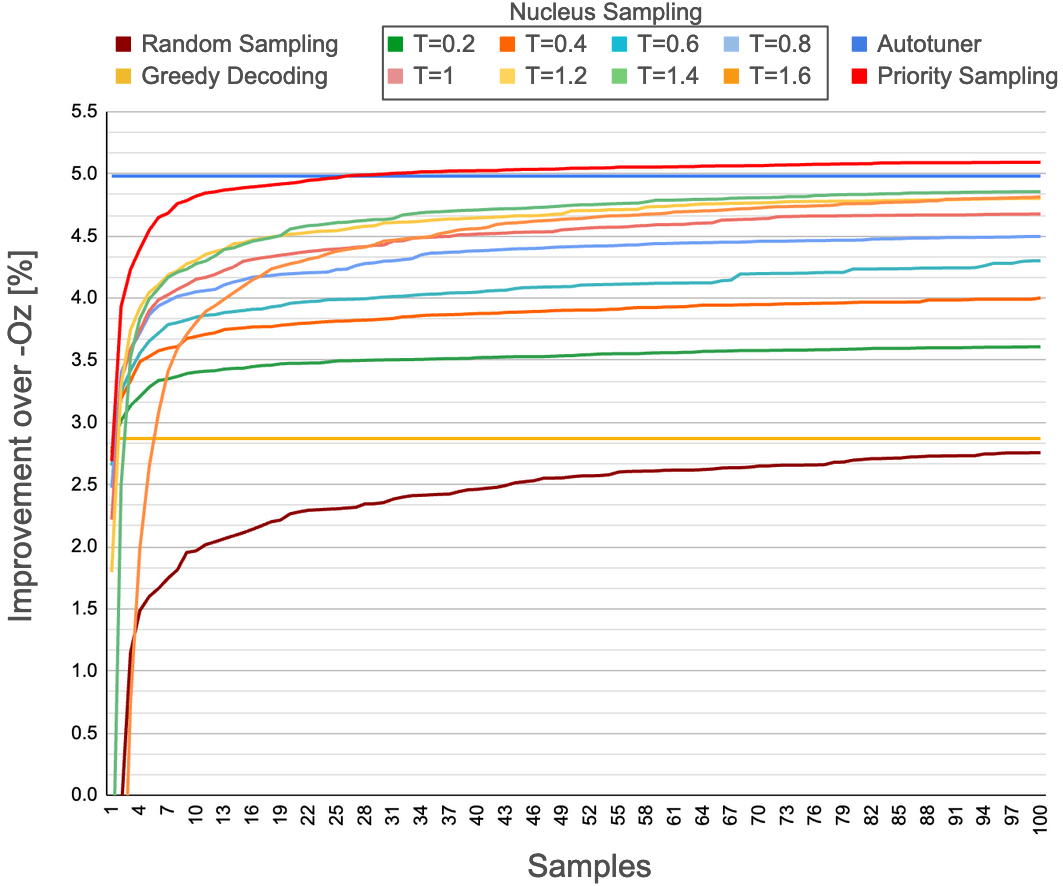}
  \caption{Average improvement in code size over -Oz optimization on 50k unseen test examples. Autotuner spends 760s for optimizing each example and sets the labels for LLM fine-tuning~\cite{cummins2023large}. Greedy Decoding, Nucleus Sampling, and Priority Sampling utilize the fine-tuned model. Random Sampling selects 100 random flags for each sample.
  Priority Sampling outperforms all previous methods including autotuner which was used for labeling.}
  \label{fig:regex_sampling_vs_temperature}
\end{figure}

We evaluate Priority Sampling on the task of optimizing LLVM optimization passes~\cite{cummins2023large} in which the model is trained to predict optimizations found by the long-running autotuner. Priority Sampling outperforms Nucleus Sampling~\cite{ravfogel2023nucleus_topp} for any number of samples, reaches 91\% of the autotuner improvement over -Oz in just 5 samples, and even outperforms the autotuner used to generate labels for finetuning the original model with 30 samples (Figure \ref{fig:regex_sampling_vs_temperature}).

\section{Priority Sampling of the LLM}
At a high level, the Priority Sampling algorithm operates by augmenting a search tree and determining which unexplored path to expand next, based on the model's highest confidence. Our idea is simple. Always focus the search towards the most interesting direction based on previous samples, rather than determining this in advance. Additionally, avoid sample repetition, which decreases sampling power.

\begin{figure}[]
  \centering
  \includegraphics[width=0.9\textwidth]{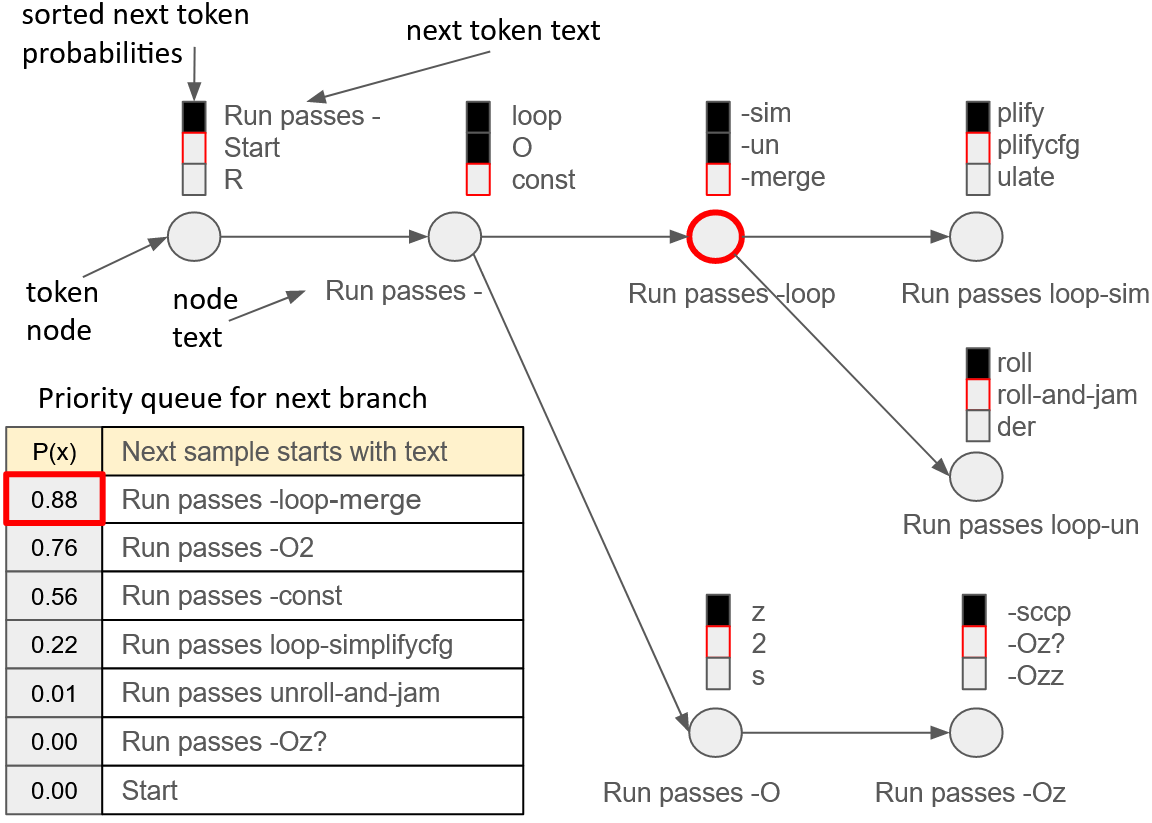}
  \caption{Priority Sampling tree expansion. Each node contains a token generated by inference and the probabilities of the next potential tokens. In the first sample, we create a branch from the root to the end-of-sequence (EOS) token and put all valid potential tokens with their probabilities in the priority queue. For every next step, branch the token that had the highest probability and generate that branch until the EOS.}
  \label{fig:unique_samples_scheme}
\end{figure}

We sketch the algorithm for Priority Sampling in Figure~\ref{fig:unique_samples_scheme}. For the first sample, the model is equivalent to Greedy Decoding, but with an important addition. With each generation, it saves top \textit{K} alternative tokens together with their predecessors in the priority queue with the token's probability as a metric. Once the sample is fully generated, we can quickly find the token prefix with the highest probability and expand the search tree from there. Since we add unexpanded tokens to the queue only once, each new sample will be unique. Additionally, we need the same number of inferences as the number of tokens generated in the search tree.

Going into more detail, the Algorithm~\ref{listing:unique_sampling} defines priority\_queue and token\_mask, which will be used for determining the best tokens prefix-sequence to expand and steer token generation to that point. Since we know the number of samples we generate, we can fix the length of priority\_queue and set token\_mask length to the generation length of the model.

With two \textit{for} loops we iterate through sample space and token generation for each sample while keeping track of the previously generated tokens for a given sample. To determine the next token, we either follow the sequence from token\_mask until we come to the branching point or expand the search tree by applying the inference. 

With inference, we get the probability distribution of tokens, from which we choose the \textit{K} tokens with the highest probabilities. An important addition here is that we exclude all tokens that don't satisfy the regular expression we define when combined with previous tokens. This can be done in constant time by using a finite state machine as described in the previous work ~\cite{willard2023regex_sampling}. This technique enables us to steer generation only towards legal format which is particularly useful for code generation.

Once we select the best tokens, we expand the search tree directly with the best token, while putting the rest of the tokens on the queue. We repeat this until we finalize the generation of the current sample. After all potential tokens for expansions were saved to the queue, we update \textit{token\_mask} with the token prefix with the highest probability. Finally, we use the token prefix to locate the node that needs to be expanded and start the generation of the new sample from there.

Priority Sampling has the algorithmic complexity of \textit{O(T*(inference + Klog(V)))}, where T is the number of generated tokens, K is the number of top-k samples we consider and V is vocabulary size. In practice, this is similar to Nucleus Sampling since the cost of inference is much higher than \textit{Klog(V)} and Priority Sampling reuses the inferences for the samples with the same prefix. 

Additionally, the memory requirements are significantly reduced by keeping the size of the priority queue constant, equal to the number of samples we generate. This way we avoid saving the probabilities of all tokens in the vocabulary for each node in the search tree while ensuring that there will be enough candidates for branching the search tree.

\begin{algorithm}[h]
\caption{Priority Sampling}
\label{listing:unique_sampling}
\begin{algorithmic}[1]
\State $priority\_queue \gets queue()$
\State $token\_mask \gets list()$
\State $sample\_tokens \gets list()$
\For{$sample: \text{range(samples)}$}
    \State $generated\_tokens = list()$
    \For{$pos: \text{range(generation\_length)}$}
        \If{$pos < \text{len(token\_mask)}$}
            \State $next\_token \gets \text{token\_mask[pos]}$
        \Else
            \State $logits \gets inference(generated\_tokens)$
            \State $best\_tokens \gets \text{choose\_best\_tokens(logits, generated\_tokens, regex)}$
            \State $next\_probability, next\_token \gets best\_tokens[0]$
            
            \For{$probability, token: \text{best\_tokens[1:]}$}
                \State $priority\_queue.push(probability, generated\_tokens + [token])$
            \EndFor 
        \EndIf
        
        \State $generated\_tokens.append(next\_token)$
    \EndFor
    \State $sample\_tokens.append(generated\_tokens)$
    \State $token\_mask \gets priority\_queue.pop()$
\EndFor
\State \textbf{return} $sample\_tokens$
\end{algorithmic}
\end{algorithm}

\newpage

\section{Evaluation}
We evaluate the Priority Sampling technique on the task of generating efficient LLVM optimization passes with LLM that reduce code size~\cite{cummins2023large}. First, we train the 7B parameter model with Llama2 architecture for 30,000 steps on 64 V100s for a total training time of 620 GPU days. The training dataset consists of 1M LLVM IR labeled with the LLVM optimization sequence found by autotuner. To generate a label for each example, the autotuner spends 13 minutes exploring 37,424 optimization passes on average. Finally, we autotune 50k unseen test examples for 13 minutes for a total improvement of 4.98\% over -Oz. 

To evaluate the effectiveness of the Priority Sampling, we compare it to the Random Sampling, Greedy Decoding, and the Nucleus Sampling for 100 steps. Random Sampling evaluates random 100 optimization passes and for each sample calculates the best optimization pass so far. Greedy Decoding generates an optimization sequence by deterministically predicting the next token with the highest probability.
For the Nucleus Sampling, we evaluate the model for temperature in the range \{0.2, 0.4, 0.6, 0.8, 1, 1.2, 1.4, 1.6\}. We found that for our problem and model architecture, temperature 1.2 is the most effective under 20 samples, while temperature 1.4 is the most effective for more than 20 samples.

We present the comparisons in Figure \ref{fig:regex_sampling_vs_temperature}. Priority Sampling outperforms Random Sampling, Greedy Decoding, and Nucleus Sampling for any number of samples. Moreover, Priority Sampling is much more sample efficient than Nucleus Sampling achieving even the performance of the autotuner in 30 steps.  Increasing the performance of the original model from 2.87\% to over 5\% with just 30 samples means that a significant part of knowledge is accessible by expanding the search tree wisely. 

This is an astonishing result since the autotuner was trained to mimic the behavior of the autotuner, and not to outperform it. Since the autotuner operates on the complex set of LLVM optimizations tied to the input program's structure, our model seems to generalize from these patterns and combines them in a novel way on the unseen programs, which results in higher performance.

\newpage

\section{Ablations}

For ablation (Table \ref{tab:results}) we show how the performance of the program changes if:
\begin{enumerate}
    \item We don't use regular expression filtering, 
    \vspace{0.1in}
    \item We use the geometric mean of probabilities of previously generated tokens as the metric for the priority queue, 
    \vspace{0.1in}
    \item We constrain the expansion for each node to 3 and 5. 
\end{enumerate}

\begin{table*}
  \centering
  \begin{tabular}{|c|c|c|c|c|c|c|}
    \hline
    & \multicolumn{5}{c}{Improvement over -Oz [\%]} &\\
    \hline
    Method & Sample 1 & Sample 3 & Sample 5 & Sample 10 & Sample 30 & Sample 100 \\
    \hline
    Random Sampling & -12.56\% & 1.15\% & 1.60\% & 1.97\% & 2.36\% & 2.76\% \\
    Temp0 & 2.87\% & - & - & - & - & - \\
    Temp1.2 & 1.80\% & 3.74\% & 4.05\% & 4.31\% & 4.61\% & 4.80\% \\
    Temp1.4 & -1.19\% & 3.52\% & 3.99\% & 4.28\% & 4.63\% & 4.86\% \\
    Temp1.6 & -10.06\% & 0.75\% & 2.65\% & 3.81\% & 4.46\% & 4.82\% \\
    \hline
    Autotuner & \multicolumn{5}{c}{4.98\%} & \\
    \hline
    Priority Sampling (PS) & 2.69\% & \textbf{4.23\%} & 4.55\% & 4.82\% & \textbf{5.00\%} & 5.09\% \\
    \hline
    PS (no regex) & \textbf{3.17\%} & 4.18\% & 4.41\% & 4.64\% & 4.93\% & \textbf{5.12}\% \\
    PS (max\_branch 3) & 2.62\% & 4.22\% & 4.56\% & 4.83\% & 4.99\% & 5.09\% \\
    PS (max\_branch 5) & 2.62\% & 4.22\% & \textbf{4.61\%} & \textbf{4.85\%} & 4.99\% & 5.09\% \\
    PS geometric (PSG) & 2.68\% & 4.17\% & 4.45\% & 4.75\% & 4.96\% & 5.07\% \\
    PSG (max\_branch 3)  & 2.62\% & 4.17\% & 4.52\% & 4.77\% & 4.98\% & 5.11\% \\
    PSG (max\_branch 5)  & 2.62\% & 4.17\% & 4.56\% & 4.80\% & 4.98\% & \textbf{5.12\%} \\
    \hline
  \end{tabular}
  \caption{Experimental results and ablation experiments of Priority Sampling. Evaluation includes the improvement of Random Sampling, Nucleus Sampling, and Autotuner over the compiler (default -Oz optimization). Ablation evaluates the use of the regular expression, constraining branching factor, and using the geometric mean as the priority metric in Priority Sampling. }
  \label{tab:results}
\end{table*}

If we don't enforce regular expression generation, the generated sampling tree will have higher probabilities, but generated samples could lead to invalid generation. To address this, we apply an additional pass that removes all invalid optimization passes and defaults to -Oz if all passes are illegal. For 1 and 100 samples this technique is beneficial, while enforcing regular expressions outperforms slightly non-constrained versions otherwise. 

Next, we evaluate using the geometric mean as the metric for the priority\_queue. This could be an interesting idea since the probability of the next token is highly biased with few previous tokens. For example prefix \textit{-mem2} will put high probability to token \textit{reg}, independently if \textit{-mem2reg} is a good optimization to apply. On the other hand, calculating geometric mean doubles memory requirements since we need to store probability with each generated token. We found that this doesn't have a significant influence on the final performance.

Finally, we evaluate the performance of the method when the branching size is constraining. This idea focuses on increasing sample diversity and avoiding the generation of many nodes with the same prefixes. For a given prefix, the first few samples should be enough to finalize the optimization strategy, while we should use other samples to explore the alternatives. Our results suggest that there is some benefit of constraining the branching factor to 5 for our problem, but not significant.

\section{Related Work}

\textbf{Stochastic Methods} introduce noise in the process of selecting the next token, resulting in increasing diversity of generation. Top-k Sampling narrows the choice of the next token to the top k most probable tokens~\cite{fan2018hierarchical}. Nucleus Sampling~\cite{holtzman2019curious} eliminates the low-probability tail of the distribution and preserves diversity by sampling from tokens whose sum is larger than \textit{top-p=0.95} probability. Mirostat~\cite{basu2020mirostat} provides a mechanism for controlling the perplexity of the generated text. Noisy Parallel Approximate Decoding~\cite{cho2016noisy} inserts unstructured Gaussian noise in each layer resulting in diverse samples. Unlike Stochastic Methods, Priority Sampling guarantees diverse set of samples deterministically.

\textbf{Beam Methods} manipulate the expansion mechanisms by constructing a search tree and focusing the direction of exploration. Diverse Beam Search ~\cite{vijayakumar2016diverse} decodes diverse lists by dividing the beam budget into groups and enforcing diversity between groups of beams. Determinantal beam search ~\cite{meister2021determinantal} defines beam search as an iterative subdeterminant maximization problem and encodes the diversity as an optimization metric. Conditional Poisson Stochastic Beam Search~\cite{meister2022typical} sample K candidates without replacement according to the conditional Poisson sampling design, resulting in a low-variance consistent estimator.

Instead of shaping reward function based on diversity, Stochastic Beam Search~\cite{kool2019stochastic} uses Gumbel-Max trick~\cite{gumbel1954statistical} to sample top-k tokens with the highest probabilities in differentiable manner. The Gumbel-Max Trick involves adding Gumbel-distributed noise to the logits (likelihood scores), after which we apply the softmax and select top-k candidates. By applying this procedure recursively for each next token, Stochastic Beam Search returns a fixed-size batch of samples.

Although Stochastic Beam Search and Gumbel top-k sampling guarantee a different output for each beam element, they are not easy to parallelize. Arithmetic Sampling~\cite{vilnis2023arithmetic} solves this problem by first sampling N numbers from uniform(0, 1) distribution and then recursively expanding tokens whose probability interval includes any of the given numbers. This method guarantees a diverse set of samples with high probability which is easy to parallelize, although it may include duplicates.

Unique Randomizer~\cite{shi2020incremental} incrementally samples sequence models while guaranteeing the uniqueness of each sample. Unique Randomizer constructs a trie to keep track of probability distribution mass for each token. Every time a sample is fully generated it subtracts it's probability from parent nodes, guaranteeing that the sample will not be selected in the future.

In our work, we extend further the idea of a Unique Randomizer with few key differences. First, we augment the trie of generated tokens with a priority queue that keeps the probabilities of each token together with its prefix. This enables us to quickly and deterministically find the node in the trie that needs to be expanded next while avoiding inferences for the prefix tokens. Additionally, we expand the priority queue with tokens that together with prefixes satisfy the regular expression that we provide, keeping the size of the trie minimal.

\textbf{Controlable text generation.} Zhang et. al ~\cite{zhang2023survey} describes a comprehensive list of challenges for controllable text generation. Unlikelyhood training~\cite{welleck2019neural} introduces a novel training objective that explicitly decreases the probability of unlikely generations. Lagutin et. al ~\cite{lagutin2021implicit} proposed Implicit Unlikelyhood Training which uses policy gradient reinforcement learning to reduce the repetition of generated text. Willard et. al~\cite{willard2023regex_sampling} introduced an efficient guiding algorithm for guiding inference of LLMs based on regular expressions. This research direction is orthogonal to our approach and could be used together with Priority Sampling.

\section{Limitations}
Priority Sampling provides an efficient way to get high-quality diverse samples, but it comes with a few limitations. The current implementation is inherently sequential. To decide what branch needs to be expanded next, it needs to construct an augmented search tree. One way to parallelize Priority Sampling would be to treat the priority queue as a task generator, from which threads take the next branching position whenever they are idle. Second, Priority Sampling needs to find the top N next tokens that match the regular expression, which is more time-consuming than sampling methods such as Unique Randomizer~\cite{shi2020incremental} and Arithmetic Sampling~\cite{vilnis2023arithmetic}. This is however necessary step for generating samples in order.

\section{Conclusion}
We present Priority Sampling, a simple inference technique that provides a deterministic and controllable way of generating unique samples for which LLM is the most confident. Priority Sampling is much more sample efficient than widely used Nucleus Sampling and outperforms it for any number of samples. 

We evaluate our model on an LLVM pass ordering task, in which our model is trained to predict the optimization passes found by a long-running autotuner. Our model was able to boost the performance of the original model from 2.87\% to 5\% improvement over default optimization -Oz in 30 steps and even more to outperform the autotuner which was used to generate training labels.

This is an astonishing result that supports the argument that LLMs store a large amount of knowledge that is accessible with the clever expansion of the search tree. Additionally, Priority Sampling includes support for regular expression generation that provides a controllable and structured exploration process.

\newpage
\bibliography{ref.bib} 
\bibliographystyle{plain}


\end{document}